\documentclass{article}

\usepackage[preprint]{neurips_2025}
\makeatletter
\renewcommand{\@noticestring}{Preprint. Work in progress.}
\makeatother

\usepackage[utf8]{inputenc} 
\usepackage[T1]{fontenc}    
\usepackage{hyperref}       
\usepackage{url}            
\usepackage{booktabs}       
\usepackage{amsfonts}       
\usepackage{nicefrac}       
\usepackage{microtype}      
\usepackage{xcolor}         
\usepackage{graphicx}
\usepackage{amsmath}
\usepackage{algorithm}
\usepackage{algorithmic}
\usepackage{tabularx}  
\usepackage{multirow}
\usepackage{threeparttable} 
\usepackage{float}
\usepackage{mathabx}

\title{BACON: A fully explainable AI model with graded logic for decision making problems}

\author{%
  Haishi Bai \\
  Department of Information Systems\\
  University of Maryland, Baltimore County (UMBC)\\
  Maryland, USA \\
  \texttt{hbai1@umbc.edu} \\
  \And
  Jozo Dujmović \\
  Department of Computer Science \\
  San Francisco State University \\
  San Francisco, CA, USA \\
  \texttt{jozo@sfsu.edu} \\
  \And
  Jianwu Wang \\
  Department of Information Systems\\
  UMBC \\
  Maryland, USA \\
  \texttt{jianwu@umbc.edu} \\
}

\begin{document}

\maketitle

\begin{abstract}
As machine learning models and autonomous agents are increasingly deployed in high-stakes, real-world domains such as healthcare, security, finance, and robotics, the need for transparent and trustworthy explanations has become critical. To ensure end-to-end transparency of AI decisions, we need models that are not only accurate but also fully explainable and human-tunable. We introduce BACON, a novel framework for automatically training explainable AI models for decision making problems using graded logic. BACON achieves high predictive accuracy while offering full structural transparency and precise, logic-based symbolic explanations, enabling effective human-AI collaboration and expert-guided refinement. We evaluate BACON with a diverse set of scenarios: classic Boolean approximation, Iris flower classification, house purchasing decisions and breast cancer diagnosis. In each case, BACON provides high-performance models while producing compact, human-verifiable decision logic. These results demonstrate BACON's potential as a practical and principled approach for delivering crisp, trustworthy explainable AI\footnote{Our code is available at: https://anonymous.4open.science/r/bacon-net}.
\end{abstract}

\section{Introduction}
With the rapid advancement of generative AI, humanoid robotics, and autonomous agents, we are entering a new era where human-AI collaboration becomes the norm. A major challenge to this collaboration lies in the probabilistic and opaque nature of modern AI models. As AI-driven agents begin to take proactive actions on behalf of humans—especially in life-threatening or mission-critical contexts—high precision alone is not sufficient. We must also ensure full transparency in how decisions are made. Only with end-to-end explainability can we establish grounded trust between humans and machines, paving the way for more responsible and effective integration of AI into our daily lives.

Traditional symbolic intelligence, grounded in human-crafted rules, offers clear and interpretable logic, but it is expensive to maintain, prone to human error, and struggles to adapt to change. Conversely, machine-learned models, though powerful, often function as black boxes—obscuring the reasoning process behind their predictions.

We introduce BACON, a novel AI model architecture designed to bridge this gap. A BACON model is inherently explainable, producing a graded logic aggregation tree that transparently captures feature interactions and decision logic. At the same time, it is fully trainable from data, enabling it to adapt like a modern learning system. BACON also facilitates human-AI collaboration during and after training, allowing experts to inject domain knowledge into the evolving model while still benefiting from the discovery capabilities of machine learning. The resulting BACON networks are accurate, transparent, and extremely lightweight, and can be pruned to meet resource constraints without sacrificing accuracy or interpretability. This makes BACON particularly well suited for deployment in battery-sensitive or resource-constrained environments, such as edge devices, intelligent appliances, and humanoid robots.

\subsection{End-to-end explainability}
We define an AI model as end-to-end explainable if it meets several essential criteria. This definition builds upon the foundational principles of interpretability and explainability discussed by Doshi-Velez and Kim \citep{doshi}, Lipton \citep{lipton2017mythosmodelinterpretability}, and Molnar \citep{molnar2025}, emphasizing complete process transparency, feature attribution, feature interaction modeling, and multi-level interpretability. 

The model must provide detailed \textbf{feature attribution} by quantifying the contribution of each input feature to the final decision. The model should filter out irrelevant features, assign meaningful importance to relevant ones, and clearly articulate how each contributes to the output. This criterion is critical because it forms the foundation of data-driven decision-making: understanding which inputs drive outcomes enables users to trust, audit, and refine the model in a principled way.

The model should exhibit \textbf{compositional transparency}—that is, it should clearly reveal how individual features and their combinations contribute to predictions. This includes uncovering logical interactions among features, such as conjunctive relationships where multiple features must act together (simultaneity), and disjunctive relationships where alternative features or combinations can lead to similar outcomes (replicability). Compositional transparency grounded in these logical structures is essential for interpreting how decisions are formed, verifying model behavior, and supporting human reasoning over the model’s logic.

The model must explicitly represent the \textbf{full decision-making pathway}, tracing how raw input features are progressively transformed into the final output. This end-to-end traceability should avoid reliance on opaque or uninterpretable intermediate computations, ensuring that every step in the reasoning process is accessible and understandable. Such transparency is essential for auditing, diagnosing, and aligning model behavior with domain knowledge.

The model must support both \textbf{model-wide and instance-specific explanations}. Model-wide explanations should capture the overarching logic and structural principles that govern the model’s behavior, while instance-specific explanations should account for individual predictions by tracing the specific factors that influenced a given instance. This dual-level interpretability is crucial for understanding general model behavior as well as diagnosing case-specific decisions.

The model must strive to \textbf{align with human decision-making processes}. It should generate decision structures that reflect commonsense reasoning and logical aggregation patterns familiar to human cognition. This alignment is particularly critical when models operate in domains closely tied to human life and judgment, such as decision-making humanoids functioning in social or safety-critical environments. By combining cognitive alignment with mechanical traceability, the model fosters greater interpretability, transparency, and trustworthiness, enabling the system to be effectively audited, explained, and debugged in sensitive or high-stakes contexts.
\subsection{Human-AI collaboration}
Human-AI collaboration is a key factor in building transparent and trustworthy AI systems. An end-to-end explainable model must be thoroughly examinable by human experts, with decision logic that is understandable and traceable from a human perspective. Human-in-the-loop review plays a critical role in ensuring the trustworthiness, correctness, and continual improvement of AI models. In addition, AI systems should provide mechanisms for domain experts to inject their knowledge and reasoning patterns into the model structure and behavior. Ideally, a model should also support post-training adjustments, enabling human experts to fine-tune model outputs based on additional domain knowledge or external constraints. This flexible collaboration between human reasoning and machine learning strengthens both the interpretability and the practical utility of the resulting models.

\section{Background and Related Works}
\subsection{Automated symbolic intelligence for explainable AI}
The key contribution of BACON is its ability to automatically train a symbolic, interpretable model for decision-making problems. Hybrid neural-symbolic architectures have long been advocated by AI visionaries such as Gary Marcus \citep{gary-marcus}, who emphasize the need to combine the robustness of symbolic reasoning with the adaptability of neural networks. One notable early effort is CILP, which encodes symbolic logic rules into neural networks and supports inductive learning \citep{neural-symbolic}. While systems like AI Feynman are designed for symbolic regression of mathematical laws in physical systems \citep{science.1165893}, BACON focuses on learning logic-based classifiers from real-world, noisy datasets, producing compact and transparent decision logic for high-stakes applications. 

Recent approaches such as Kolmogorov–Arnold Networks (KANs) \citep{kan} aim to enhance interpretability by replacing fixed activation functions with learnable splines. However, they still lack an explicitly symbolic decision model, making their internal reasoning opaque to users. Similarly, the Mechanistic AI Scientist Simulator (MASS) \citep{mass} explores whether AI models trained on scientific problems can converge on shared theories. MASS relies on physics-informed neural networks, which enable interpretable predictions grounded in physical laws, but do not employ symbolic logic structures or explicit reasoning trees. Neural Logic Machines (NLMs) \citep{nlm} extend neural networks toward relational reasoning using tensorized architectures, yet their outputs remain in latent vector forms, limiting their ability to produce human-readable explanations. In contrast to these approaches, BACON employs a Logic Scoring of Preference (LSP) aggregation tree\citep{lsp} that is inherently interpretable. The resulting tree not only ranks feature importance, but also captures intrinsic feature characteristics and interrelationships—such as mandatory conditions, preferred values, optional traits, and mutually exclusive attributes—providing a complete, transparent reasoning process from input features to final decision output.

\subsection{Graded Logic and the LSP decision method }

Graded Logic (GL) is a continuum-valued propositional logic obtained as a human-centric generalization of classical Boolean logic \citep{lsp, lsp2019, lsp2025}. GL is a mathematical model of human commonsense logical reasoning with graded percepts. Four fundamental graded percepts include the graded truth (logic variables \(x_i \in I = [0, 1], i=1, \ldots , n, n > 1\)), the graded importance of logic variables (\(w_i \in [0,1], i = 1, \ldots, n, \sum_{i=1}^{n} w_i=1 \)), and the graded conjunction and disjunction (models of simultaneity and  substitutability). In GL, the graded conjunction and disjunction are complementary and unified in a single logic function called Graded Conjunction/Disjunction (\(GCD_n: I^n \to I\)). Logic properties of GCD are characterized by the degree of conjunction (andness, \(\alpha\)) and the complementary degree of disjunction (orness, \(\omega = 1-\alpha\)). These parameters for the GCD logic function were introduced in \citep{lsp1974} as follows:
\begin{align*}
\alpha = 1 - \omega = \frac{n}{n - 1} - \frac{n + 1}{n - 1}
\int_0^1 \cdots \int_0^1 \text{GCD}_n(x_1, \ldots, x_n; w_1, \ldots,w_n, r) \, dx_1 \ldots dx_n, \\
w_i=1/n, i = 1, \ldots, n,  r \in \mathbb{R}, \quad -1/(n-1) \leq \alpha \leq n/(n-1).
\end{align*}

So, the andness and orness are defined assuming that all variables have the same importance, and the adjustable parameter \(r\) is used to select desired logic properties of the GCD function. For any given number of variables, the andness is a function of the GCD parameter: \(\alpha=f(r)\). Since the andness is a desired adjustable input parameter of the GCD function, we can insert \(r=f^{-1}(\alpha)\) in GCD, and such a function is denoted \(GCD(\mathbf{X};\mathbf{W}, \alpha)\) and called the andness-directed GCD. 
Here \(\mathbf{X}=(x_1,\dots,x_n)\) denotes a vector of input degrees of truth, and their relative importance is adjusted using the vector of weights \(\mathbf{W}=(w_1,\dots,w_n)\). If GCD is the pure conjunction \(min(x_1,\dots,x_n)\), then \(\alpha=1\) and \(\omega=0\).  If GCD is the pure disjunction \(max(x_1,\dots,x_n)\), then \(\alpha=0\) and \(\omega=1\). Thus, by adjusting andness in the range  \(0 \leq \alpha \leq 1\) we can realize a continuous transition from the pure disjunction to the pure conjunction. If \(\alpha=\omega=0.5\), then GCD has balanced conjunctive and disjunctive properties and becomes the logic neutrality, or the arithmetic mean \(GCD_n(\mathbf{X};\mathbf{W}, 0.5)=\sum_{i=1}^{n}w_ix_i\).

\begin{figure}[ht]
  \centering
  \includegraphics[width=.9\textwidth]{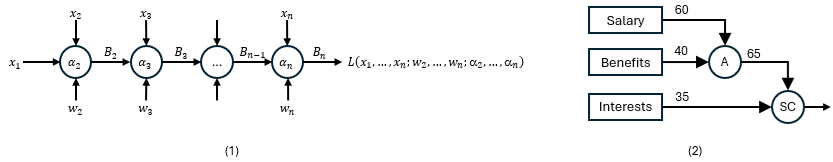}
  \caption{(1) A graded propositional logic model structured as a binary tree; (2) A sample job selection LSP tree}
  \label{fig:lsp-binary}
\end{figure}

All formulas of graded propositional calculus are obtained by superposition of GCD and negation \(not(x)=1-x\). A particularly effective regular structure of logic aggregation is the binary tree structure shown in Figure \ref{fig:lsp-binary}-1. We use the aggregation function \(L: I^n \to I\) resursively structured as a sequence of \(GCD_2\) aggregators \(B_i(x_{i-1},x_i)=GCD_2(x_{i-1},x_i;1-w_i,w_i,\alpha_i), i =2,\dots\,n\) where each \(GCD_2\) aggregator has two adjustable parameters: the andness \(\alpha_i\) and the weight \(w_i\) the “newcomer argument” \(x_i\). In this recursive model, the argument \(x_n\) is a newcomer that contributes its individual positive impact to the group impact of previous \(i-1\) arguments. The LSP method defines a family of andness-guided aggregators that model human decision-making logic. These aggregators can be composed into a complete, interpretable decision model. Figure 1-2 illustrates a job selection LSP aggregation tree, where salary and benefits are merged using a weighted neutral aggregator (A), allowing their weighted contributions to combine additively into a compensation score. This score is then aggregated with interests via a soft conjunctive aggregator (SC), which favors situations where both criteria contribute meaningfully but tolerates some imbalance, producing an interpretable job suitability score.

\section{BACON architecture}
The BACON architecture is designed from the ground up to automatically discover feature characteristics and relationships. It constructs precise, interpretable tree structures using graded logic. The resulting tree can be pruned to remove irrelevant inputs without sacrificing accuracy, and its internal logic can be extracted as a simple arithmetic function that can be evaluated independently of AI frameworks. Additionally, the aggregation logic can be translated into human-readable reports (e.g., using an LLM), as shown in Figure \ref{fig:architecture} (see Appendix \ref{appendix:h} for a sample prompt and report).

\begin{figure}[ht]
  \centering
  \includegraphics[width=\textwidth]{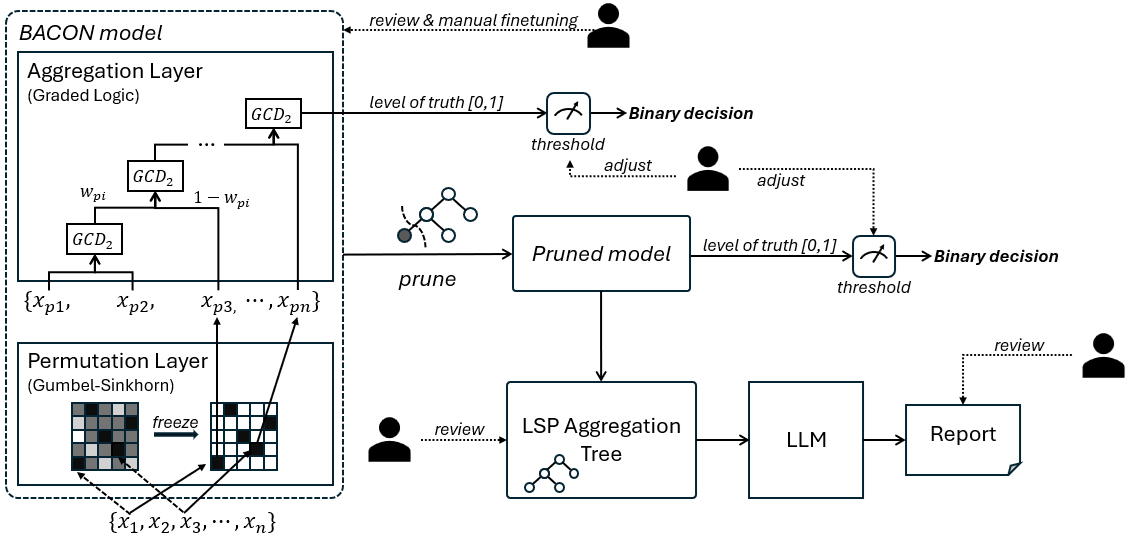}
  \caption{High-level architecture of BACON}
  \label{fig:architecture}
\end{figure}

\subsection{BACON model architecture}

A BACON model consists of two layers: an input permutation layer based on the Gumbel-Sinkhorn network \citep{sinkhorn} and a binary tree aggregation layer. The input permutation layer enables the model to explore different input orderings, while the binary tree layer constructs a binary structure that can be translated into an LSP aggregation tree (see Figure \ref{fig:lsp-binary}). The GCD operator in BACON replaces the traditional weighted sum used in standard neurons, effectively defining a new type of neuron based on logic aggregation rather than linear combination. The GCD operator represents a generalized conjunction-disjunction function parameterized (weight \(w\) and andness \(\alpha\)) to express logical aggregation between two weighted inputs:
\small{
\[
z =
\begin{cases}
GCD_2(x,y,w,a) =
\begin{cases}
\text{1, if x=y=1, or 0 otherwise} & \alpha =2 \\
(x^{2w}y^{2(1-w)})^{\sqrt{\frac{3}{2 - \alpha}} - 1}, & 0.75 \leq \alpha < 2 \\
(3 - 4\alpha)(wx + (1-w)y) + \\ (4\alpha - 2)(x^{2w}y^{2(1-w)})^{\sqrt{\frac{3}{2 - \alpha}} - 1}, & 0.5 < \alpha < 0.75 \\
wx + (1-w)y, & \alpha = 0.5
\end{cases} \\
1 - GCD_2(1-x, 1-y, w, 1-\alpha), \quad -1 \leq \alpha < 0.5
\end{cases} \\
\\
\]
}

\subsection{Training process}
BACON models are trained using a custom two-phase process. In the early phase, the permutation layer is configured with a high Gumbel noise scale and Sinkhorn temperature to encourage exploration of input permutations. This allows the aggregation layer to be trained based on soft feature assignments, which are adjusted simultaneously. Once the model achieves a sufficiently low loss, the Hungarian algorithm \citep{Hungarian} is applied to the soft assignment matrix to extract a discrete permutation. If this permutation yields an acceptable loss, the model freezes the permutation layer by replacing the Gumbel-Sinkhorn soft assignment with a hard assignment matrix \(P_{hard}\). Training then continues with the aggregation layer only, allowing fine-tuning of the weights and the GCD aggregation parameters (see Algorithm \ref{bacon-training} in Appendix \ref{appendix:a}). The core technical challenge BACON faces is discovering and training an optimal feature permutation within a vast search space of size \(n!\).  In the breast cancer case, with 30 input features, this amounts to \(30! = 2.65 \times 10^{32}\)
possible permutations—an astronomically large space that cannot be exhaustively searched. By leveraging Gumbel-Sinkhorn for stochastic exploration and the Hungarian algorithm for deterministic freezing, BACON can efficiently identify high-quality permutations under given constraints. 
The loss function is calculated as: 

\[
\mathcal{L}_{\text{BACON}} = a \cdot \left( 
\varepsilon\left( 
\text{GCD}_2\left( \ldots \text{GCD}_2\left( \text{GCD}_2(x_{p1}, x_{p2}), x_{p3} \right), \ldots, x_{pn} \right), y 
\right) 
+ \lambda \cdot \frac{1}{N} \sum_{i=1}^{N} \left( \sigma(w_i) - 0.5 \right)^2 
\right)
\]

The loss function \(\mathcal{L}_{\text{BACON}}\) consists of two terms: a prediction error and a weight regularization term, both scaled by the amplification factor \(a\). The prediction error \(\varepsilon(\cdot,y)\) measures the discrepancy between the output of the generalized conjunctive/disjunctive operator \(GCD_2\), recursively applied over a permutation of input features \(x_{p1}, x_{p2}, \dots, x_{pn}\), and the ground truth \(y\). The weight regularization term penalizes deviation of each aggregator weight \(w_i\) from 0.5 by applying a sigmoid activation \(\sigma(w_i)\) and computing the squared difference from 0.5, averaged over all \(N\) nodes, with strength controlled by the coefficient \(\lambda\). This formulation promotes balanced aggregations while ensuring predictive accuracy, and supports simultaneous training of weights, andness, and feature permutations.
\subsection{How BACON delivers end-to-end explainability}
BACON is specifically designed to provide end-to-end explainability for decision-making problems. Table~\ref{tab:explainability-comparison} compares BACON with commonly used explainable models and methods across five key explainability criteria. As shown in the table, BACON matches or surpasses the explainability capabilities of other interpretable models in all categories. Its unique combination of symbolic reasoning, compositional transparency, and cognitive alignment makes BACON particularly well-suited for domains that demand transparent and trustworthy decision-making. More detailed comparisons are presented in the following sections.

\begin{table}[h]
\centering
\caption{Comparison of Explainability Across Methods}
\label{tab:explainability-comparison}
\resizebox{\textwidth}{!}{%
\begin{threeparttable}
\begin{tabular}{lcccccc}
\toprule
\textbf{Criterion} & \textbf{BACON} & \textbf{Decision Tree} & \textbf{SHAP} & \textbf{LIME} & \textbf{KAN} & \textbf{EBM} \\
\midrule
Feature attribution & \textbf{Yes} (explicit) & Yes (structure) & Yes (importance)\tnote{1} & Yes (importance)\tnote{1} & Yes (functions) & Yes (effects) \\
Compositional transparency         & \textbf{Yes}                 & Yes                 & No\tnote{2} & No\tnote{2} & Yes             & Yes \\
Full decision-making pathway       & \textbf{Yes}                 & Yes                 & No            & No            & Yes             & Yes \\
Model-wide and instance-specific explanation       & \textbf{Both}                & Both                & Local\tnote{3} & Local\tnote{3} & Both            & Both \\
Alignment with human decision-making\tnote{5}     & \textbf{Yes}                 & No\tnote{4} & No            & No        & No             & No \\
\bottomrule
\end{tabular}
\begin{tablenotes}
\footnotesize
\item[1] SHAP and LIME provide local importance scores but do not reveal how features are combined.
\item[2] Only the explanation output is compositional; the model remains opaque.
\item[3] Global understanding is possible only via aggregation of local outputs.
\item[4] Assumes the tree is small and well-pruned.
\item[5] Only BACON explicitly aims at alignment with human decision-making process.
\end{tablenotes}
\end{threeparttable}
}
\end{table}

\subsubsection{Feature attribution}
Each BACON neuron consists of two parameters: importance (\(w\)) and andness (\(\alpha\)). Because each feature is considered one at a time, it's straightforward to assess its impact by examining its associated weight and how it is aggregated into the binary tree through the corresponding andness value. For example, \(\alpha\ge11/14\)
indicates a hard conjunction, making the feature mandatory for the final decision. Similarly, \(\alpha\in[8/14,10/14]\))represents a soft conjunction, describing desired but non-mandatory features; \(\alpha\in[4/14,6/14]\)) corresponds to a soft disjunction, modeling optional features; and \(\alpha\in[-1,3/14]\)) reflects a hard disjunction, treating the feature as sufficient by itself. These behaviors mirror common decision logic patterns, allowing BACON to represent feature roles such as mandatory, desired, optional, and sufficient in an interpretable way. Further, due to the left-associative tree structure, when the importance parameter \(w\) is heavily regulated toward neutral \(0.5\), a feature needs to appear near the top of the tree to maintain significant impact, as deeper nodes naturally accumulate less weight. This enables BACON to prioritize important features toward the top, forming an implicit feature importance ranking, while progressively pushing less relevant features to the bottom where they can be trimmed.

Unlike LIME \citep{lime} and SHAP \citep{shap}, which focus on post-hoc feature attribution through approximations, BACON embeds feature importance directly into the model structure and logic, offering deterministic and transparent attribution without relying on perturbations or surrogate models. Compared to Decision Trees \cite{decision-tree}, BACON offers explicit logic quantification through andness parameters, enabling nuanced characterization of feature roles (mandatory, optional, sufficient), while decision trees typically only split features without expressing their aggregation logic.

\begin{figure}[H]
  \centering
  \includegraphics[width=\textwidth]{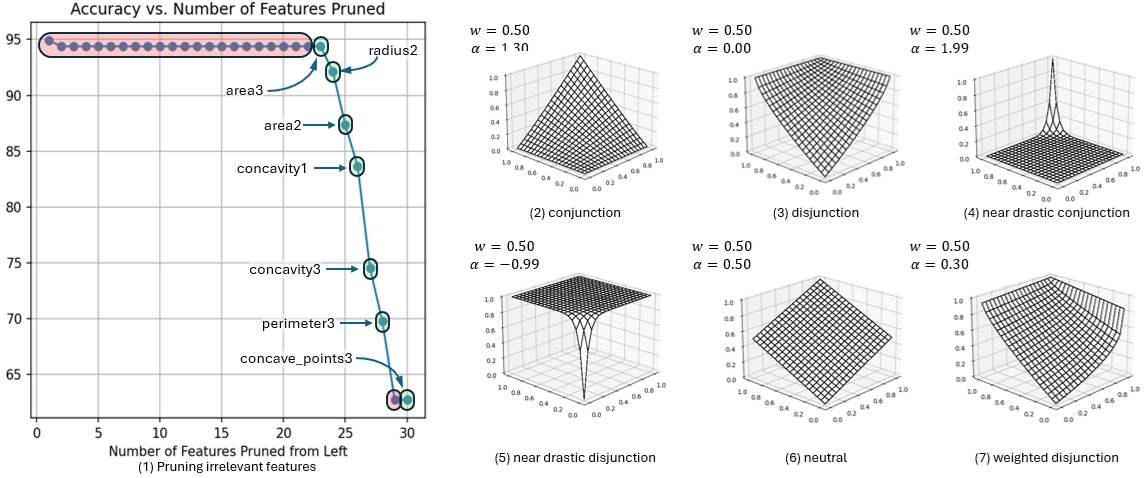}
  \caption{Feature attribution with BACON: (1) important vs. irrelevant features; (2) conjunction; (3) disjunction; (4) near drastic conjunction; (5) near drastic disjunction; (6) neutral; (7) weighted disjunction}
  \label{fig:feature-attribution}
\end{figure}

Figure~\ref{fig:feature-attribution}-1 illustrates how model accuracy is affected when tree nodes are progressively trimmed. It is evident that the most important features are positioned at the top (right side) of the tree, as pruning less relevant features has little impact on accuracy. This demonstrates the model's inherent feature prioritization mechanism. Figures~\ref{fig:feature-attribution}-2 through~\ref{fig:feature-attribution}-7 visualize how pairs of inputs are aggregated using GCD operators with varying degrees of andness. Unlike standard neural networks that rely on simple weighted summation, these operators enable more expressive logical interactions between features, such as conjunction, disjunction, and their extreme forms, thereby capturing richer decision patterns.

\subsubsection{Compositional transparency and full decision-making pathway}
The training result of the BACON network is a symbolic expression that follows the LSP model of reasoning. This provides both global model-level and local instance-level explainability. BACON distinguishes itself by offering a complete, traceable decision pathway where each decision can be followed from input to output in symbolic form. LIME and SHAP provide local explanations but do not reconstruct the full model reasoning process. Decision trees do offer transparency, but they often suffer from fragmentation and do not capture feature interactions explicitly as conjunctive or disjunctive forms. KAN aims for symbolic-like reasoning through neural interpolator but lacks BACON's explicit, human-readable aggregation logic. Moreover, KAN is primarily designed for general function approximation rather than decision-making tasks, and does not natively provide symbolic decision pathways or logical operators like BACON's LSP aggregators. EBM (Explainable Boosting Machines) \citep{ebm} provides globally consistent additive models but cannot model complex logical dependencies between features.

While KAN provides a flexible framework for learning complex functions through neural interpolators, it differs from BACON’s explicit symbolic aggregation. KAN lacks built-in logical operators like BACON’s LSP aggregators and does not natively produce traceable decision pathways. Although LSP-like knowledge could, in principle, be injected into KAN via architectural constraints or regularization, doing so is non-trivial—especially since the type of aggregation (e.g., conjunctive or disjunctive) is controlled by learned andness parameters that cannot be predetermined. Consequently, BACON’s structure remains better suited for tasks requiring explicit, human-readable logic reasoning.

\subsubsection{Full alignment with human decision-making process}
An input to a BACON network is normalized to the \([0,1]\) range, representing the degree of truth of a given statement. For example, a normalized input for cell size can be interpreted as the truth level of the statement "the cell is big." The BACON network then aggregates these truth levels into a global truth value representing the final decision. BACON adopts the LSP model for its rich modeling capabilities, its strong emphasis on alignment with human reasoning, and its proven track record over 50 years of practical use in decision-making problems. This provides BACON with unique capabilities and a focus on logical coherence that are not typically found in other explainability methods. 

A key assumption of BACON is the existence of intrinsic logical structures within the data. The network is designed to autonomously extract these structures through its training process. The BACON's training process is strictly regularized by the LSP framework, preventing the model from arbitrarily adjusting parameters solely to optimize predictive accuracy. Consequently, patterns identified by BACON inherently exhibit strong logical consistency, distinguishing it from methods that rely primarily on statistical fitting or post-hoc explanation. This also means when BACON network does not converge, it provides a strong hint that a human-interpretable logic likely does not exist, or is not easily extractable from the data.

\section{BACON in breast cancer diagnosis use case}
To evaluate BACON’s correctness, performance, and explainability, we selected a set of representative scenarios spanning diverse data types and complexities: discovering classic Boolean expressions, making house purchase decisions, classifying Iris flowers and diagnosing breast cancer. In this section, we present the details of the breast cancer diagnosis scenario, while additional information on the other scenarios is provided in the Appendices.

\subsection{Breast cancer diagnosis}
For breast cancer diagnosis experiment, we use the WDBC \citep{breast_cancer_wisconsin_(diagnostic)_17} dataset (CC-BY 4.0). The dataset contains 569 samples with 357 benign cases and 212 malignant cases. The WDBC dataset has been extensively used to benchmark black-box models, achieving impressive results such as 99.89\% accuracy with ensemble machine learning methods \citep{Reshan2023Enhancing}. However, these models are not only unexplainable but also require additional data processing steps such as feature selection or fine-tuning specific to the problem context. In contrast, we trained a BACON network using all 30 features without any pre-processing (other than normalization), feature selection, or problem-specific fine-tuning. The resulting network achieves 98.07\% (95\% CI: [97.85\%, 98.29\%]) accuracy out of the box while remaining tunable and fully explainable. Although this performance is comparable to highly optimized black-box methods, BACON additionally provides end-to-end explainability for its decision-making process. In the following sections, we examine how BACON satisfies the criteria for end-to-end explainability as defined in Section 1.1.

\subsection{Feature attribution}
Because BACON’s recursive binary tree structure considers one feature at a time, the trained tree can be progressively pruned to evaluate the contribution of each input feature in isolation. By analyzing the accuracy drop after each pruning step, BACON is able to rank features by their importance and effectively filter out irrelevant ones. Unlike conventional feature importance measures based on statistical contribution or weight magnitude, BACON ranks features by their logical impact on the model’s decision process. This approach evaluates how each feature’s presence or absence influences the overall reasoning outcome, providing a more decision-centric and interpretable ranking that reflects the feature’s role within the model’s logic. In the breast cancer case, BACON identified that 7 out of 30 features contribute meaningfully to the decision-making process, in descending order of importance: \textit{concave\_points3}, \textit{concavity1}, \textit{perimeter3}, \textit{concavity3}, \textit{radius2}, \textit{area2},  and \textit{area3} (see Figure \ref{fig:feature-attribution}-1). The remaining 23 features were found to have almost no measurable contribution to the final result and were thus considered irrelevant. The selected features demonstrate clear alignment with clinical expectations: malignant cells tend to be larger (reflected by high \textit{area}, \textit{radius} or \textit{perimeter}) and exhibit shape irregularities (high \textit{concave\_points} or \textit{concavity}) \citep{baba2007tumor}. This highlights BACON's strength in producing precise, interpretable models while preserving high predictive performance without prior medical knowledge.

\subsection{Feature aggregation}

\begin{figure}[ht]
  \centering
  \includegraphics[width=0.9\textwidth]{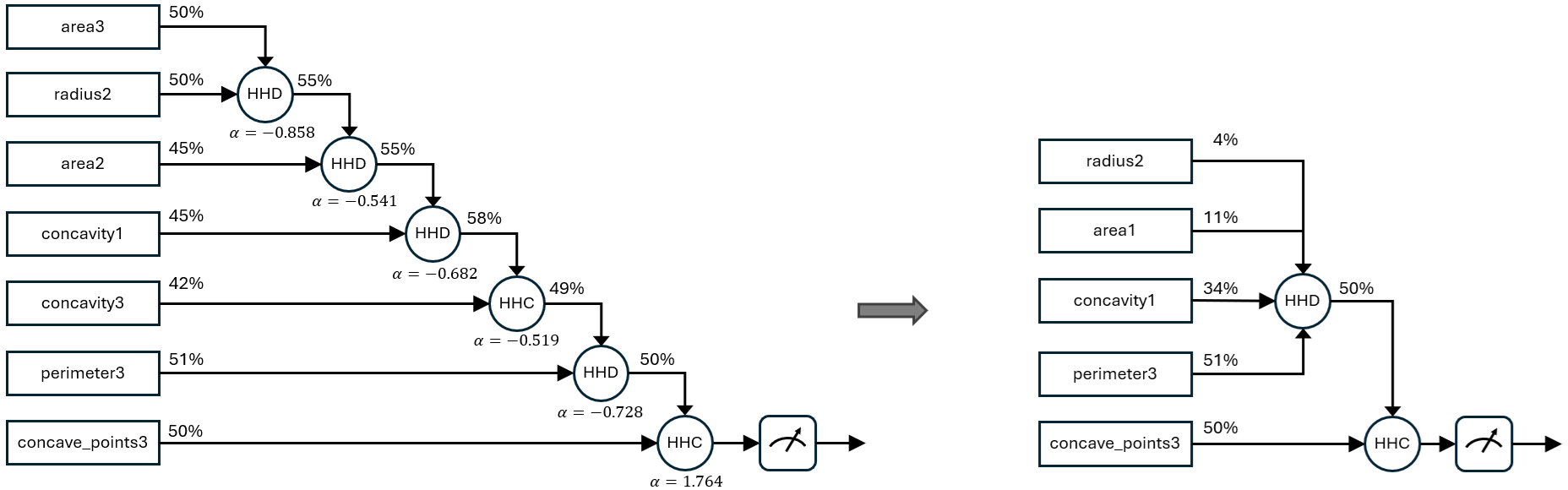}
  \caption{Breast cancer diagnosis process modeled as LSP aggregation tree}
  \label{fig:feature-aggregation}
\end{figure}

Figure~\ref{fig:feature-aggregation} (left) presents the LSP aggregation tree constructed from the pruned BACON network, illustrating how the model composes its decision from the bottom up. Due to high correlations among certain features (such as between \textit{concavity1} and \textit{concavity3}), the LSP tree can be further simplified to offer a more concise reasoning process for human interpretation by combining features with similar semantic identities (such as \textit{area} features and \textit{concavity} features), as shown in Figure~\ref{fig:feature-aggregation} (right).

The LSP aggregation reveals a clear and interpretable decision path. First, the top four features are combined using a disjunctive aggregator, forming a "size group" that characterizes a large cell when any of these features exhibit high values. This group is then combined with \textit{concavity\_points3} through a highly conjunctive aggregator, enforcing that a malignant cell must simultaneously be large and irregular. Notably, \textit{concavity1}, which measures the number of "dents" in a cell's contour, appears in the size group to compensate for smaller sizes, highlighting that cell irregularity plays a decisive role in the diagnosis. This observation aligns with findings from prior studies, such as \citep{Yu2023}.

\subsection{Human review and fine-tuning}

\begin{figure}[ht]
  \centering
  \includegraphics[width=0.9\textwidth]{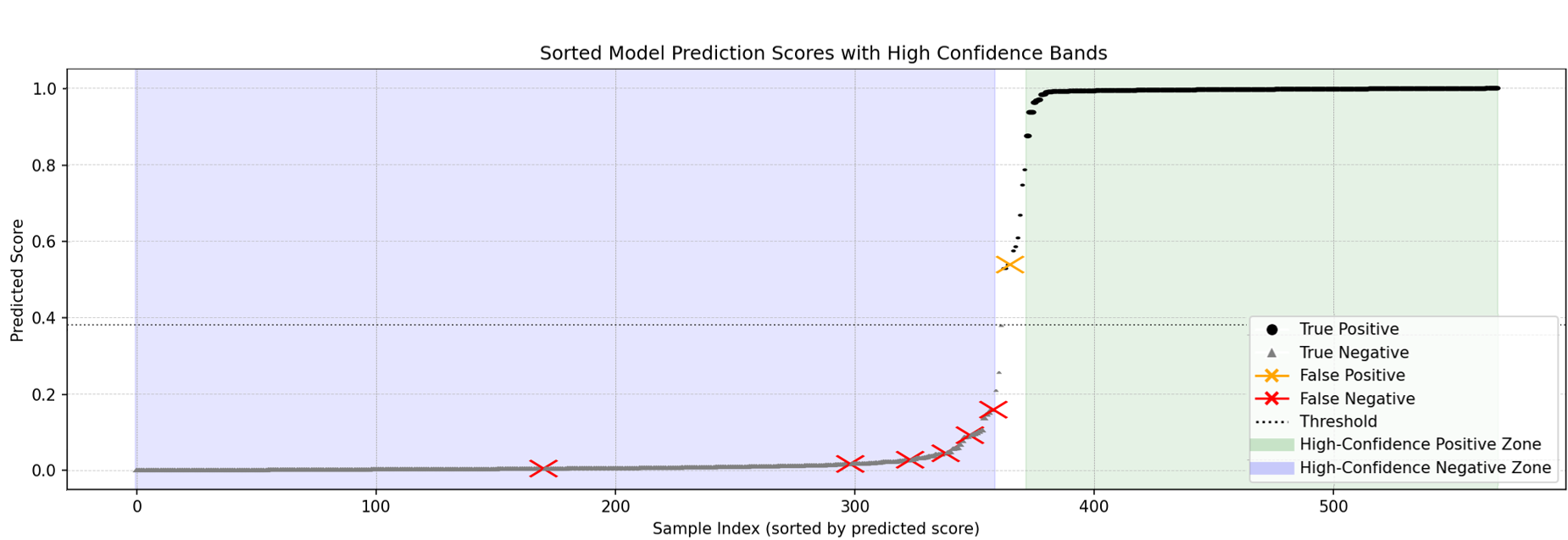}
  \caption{Converting to degree of truth to binary output (threshold = 0.5)}
  \label{fig:bacon-threshold}
\end{figure}

Before being converted to a binary output in \(\{0,1\}\), BACON's output represents a degree of truth within the range \([0,1]\). The threshold used to map this continuous value to a binary decision can be adjusted depending on the specific application context. Figure~\ref{fig:bacon-threshold} shows that BACON is generally confident in its predictions—producing outputs close to 0 for negative cases and close to 1 for positive cases. Therefore, a threshold of 0.5 is a reasonable default. However, the diagram also reveals that most misclassifications occur when the model’s output is near lower thresholds, indicating low confidence in these cases. In such situations, a human user—such as a doctor—may adjust the threshold to better suit the operational priorities. For example, during an initial screening phase, a lower threshold may be adopted to prioritize recall and reduce the risk of missing potential cases. In contrast, during diagnosis or treatment planning, a higher threshold can be used to ensure decisions are made with greater certainty. Table~\ref{tab:thresholds} summarizes the optimal threshold choices for different application scenarios.

\begin{table}
    \centering
    \caption{Human fine-tuning with different thresholds}
    \label{tab:thresholds}
    \begin{tabular}{lrrrr}
        \toprule
        \textbf{Scenario} & \textbf{Threshold} & \textbf{Accuracy}  & \textbf{Precision} &\textbf{Recall}\\ \midrule
        Initial screening & 0.01  & 63.97\%  & 50.84\% & \textbf{100.00\%} \\
        Clinical diagnosis & 0.46 & \textbf{98.77\%} & 99.52\% & 97.17\% \\
        Treatment planning  & 0.63  & 98.42\%  & \textbf{100.00\%} & 95.75\% \\        
        \bottomrule
    \end{tabular}
\end{table}

In summary, our breast cancer diagnosis experiment using the WDBC dataset demonstrated BACON’s strong interpretability and diagnostic power. Its pruning mechanism revealed that out of all 30 features, only a few key features are critical and its conclusion aligns closely with clinical understanding. This ability to simplify decision logic without sacrificing accuracy underscores BACON’s potential as a valuable tool for interpretable AI in medical diagnostics.

\section{Discussions and future studies}
Thus far, our experiments have focused on relatively small problems with up to 38 features—within the bounds of human interpretability. For more complex tasks, a common approach is to decompose the problem into subsystems, evaluate each individually, and then aggregate the results. To automate this process, a strategy is needed to group parameters effectively so that each group remains understandable and verifiable by human experts. In theory, BACON’s methodology could be applied recursively to support such decomposition. However, the custom training process makes combining multiple BACON networks non-trivial. While this is technically feasible, the current BACON implementation does not support chaining or hierarchical composition. Moreover, the left-associative tree structure used in BACON is well-suited for feature ranking but not ideal for feature grouping. A more appropriate structure for grouping may be a balanced binary tree.

The left-associative tree structure presents two key limitations: (1) each feature is considered only once, and (2) features are processed sequentially. While this structure has proven effective in many decision-making scenarios, it limits the model's capacity to capture more complex feature interactions. A hybrid structure that combines a balanced tree with a linear tree could support both ranking and equitable feature participation. Further, a tunable hyperparameter could control the degree of balancing to suit different problem types. Currently, BACON employs the LSP GCD aggregator, which provides a smooth, continuous reasoning surface over the unit hypercube. However, different logical patterns might be better approximated using alternative aggregation functions. Exploring a variety of aggregators—or even assembling an ensemble of BACON models using diverse aggregators—may yield deeper insights into specific domains. Fundamentally, the BACON network performs a guided search over a vast space of feature permutations and aggregation paths. By leveraging different aggregation strategies, BACON could be adapted to a range of tasks beyond decision logic such as cybersecurity and policy optimization. Finally, one of BACON’s unique advantages is its ability to extract compact, interpretable expressions through pruning and structural transparency. These symbolic forms can be deployed without any underlying AI framework, making BACON especially suitable for high-throughput, low-latency, and low-cost applications such as deployment on tiny edge devices, real-time control systems, robotics and time-series data processing.

\section{Conclusion}

BACON introduces a fully explainable AI architecture that combines the adaptability of modern machine learning with the symbolic transparency of graded logic. By replacing conventional neural aggregation with a GCD operator, BACON constructs a graded logic aggregation tree that models the decision-making process end-to-end. This structure is not only interpretable and prunable, but also convertible into standalone functions suitable for deployment in edge devices and real-time environments.

Our experiments on medical diagnosis demonstrate BACON’s ability to automatically uncover interpretable, clinically aligned decision logic—achieving high accuracy without sacrificing transparency. BACON’s structure enables human-in-the-loop interaction, model simplification, and symbolic extraction, distinguishing it from traditional black-box approaches.

Future work includes scaling BACON to larger, more complex datasets through tree balancing or modular composition; expanding the framework to support alternative logic operators; and exploring ensemble strategies across multiple GCD formulations. With its unique blend of interpretability, adaptability, and hardware efficiency, BACON offers a promising path toward trustworthy, explainable AI.

\bibliographystyle{unsrtnat}
\bibliography{BACON_references}

\newpage
\appendix

\section{BACON training algorithm}
\label{appendix:a}

\begin{algorithm}
\caption{BACON model training process}
\label{bacon-training}
\textbf{Parameters:}\\[0.5em]
\begin{tabularx}{\linewidth}{lX lX}
\(N_{attempts}\)     & Number of attempts              & \(N_{epochs}\)     & Epochs per attempt \\
\(\gamma_{inc}\)     & Gumbel noise increase factor    & \(\gamma_{dec}\)   & Gumbel noise decrease factor \\
\(\epsilon_{min}\)   & Minimum Gumbel noise            & \(\epsilon_{max}\) & Maximum Gumbel noise \\
\(\tau_{decay}\)     & Sinkhorn temperature decay      & \(L_{freeze}\)     & Loss threshold to trigger freeze \\
\(L_{accept}\)       & Acceptance threshold for freezing & \(A_{target}\)   & Required model accuracy  \\
\end{tabularx}
\vspace{0.5em}
\begin{algorithmic}[1]
\FOR{attempt \(\gets 1\) to \(N_{attempts}\)}
    \FOR{epoch \(\gets 1\) to \(N_{epochs}\)}
        \IF{\( \text{epoch} \bmod 1000 = 0 \)}
            \STATE Decay Sinkhorn temperature by $\tau_{\text{decay}}$
        \ENDIF
        \STATE Forward pass
        \IF{gradients vanish or explode}
            \STATE \textbf{break}
        \ENDIF
        \STATE Record loss in sliding history window
        \IF{loss is not improving}
            \STATE Increase Gumbel noise (up to $\epsilon_{\text{max}}$) using $\gamma_{\text{inc}}$
        \ELSE
            \STATE Decrease Gumbel noise (down to $\epsilon_{\text{min}}$) using $\gamma_{\text{dec}}$
        \ENDIF
        \IF{not frozen \textbf{and} loss $<$ $L_{\text{freeze}}$}
            \STATE Generate a permutation using Hungarian algorithm
            \STATE Evaluate the pemutation
            \IF{best loss $<$ $L_{\text{accept}}$}
                \STATE Freeze Gumbel-Sinkhorn layer into $P_{\text{hard}}$
            \ENDIF
        \ENDIF
    \ENDFOR
    \IF{model is frozen}
        \STATE Evaluate model accuracy
        \IF{accuracy $\geq A_{\text{target}}$}
            \STATE \textbf{return best model}
        \ENDIF
    \ENDIF
\ENDFOR
\STATE \textbf{raise} exception if no model meets the target accuracy \\
\end{algorithmic}
\end{algorithm}

\section{BACON approximation of classic boolean expressions}
\label{appendix:classic-boolean}
Graded logic can be seen as an extension of classical Boolean logic. As such, BACON should be capable of learning and approximating classical Boolean relationships among input features. In our research, we evaluated BACON using a series of Boolean expressions involving up to eight variables. For each expression, we generated synthetic training datasets by repeating all possible input combinations 100 times to ensure robustness.

\begin{figure}[ht]
    \centering
    \includegraphics[width=.3\textwidth]{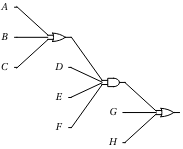}
    \caption{Boolean circuit for (((((((A or B) or C) and D) and E) and F) or G) or H) \citep{booltool}}
    \label{fig:boolean}
\end{figure}

BACON was generally able to recover the underlying Boolean logic, even for complex expressions. However, the time required to converge to the correct expression varied. In more intricate cases, it could take hours for BACON to successfully match the target logic. We also observed that BACON exhibits a bias toward conjunctions: it tends to more easily learn expressions dominated by AND operators. Expressions with deeply nested AND combinations were consistently matched more quickly and reliably than those dominated by OR or mixed logic. Figure ~\ref{fig:boolean} shows the logic structure of one such test case: 
\[
(((((((A \lor B) \lor C) \land D) \land E) \land F) \lor G) \lor H)
\]

\begin{figure}[ht]
    \centering
    \includegraphics[width=.9\textwidth]{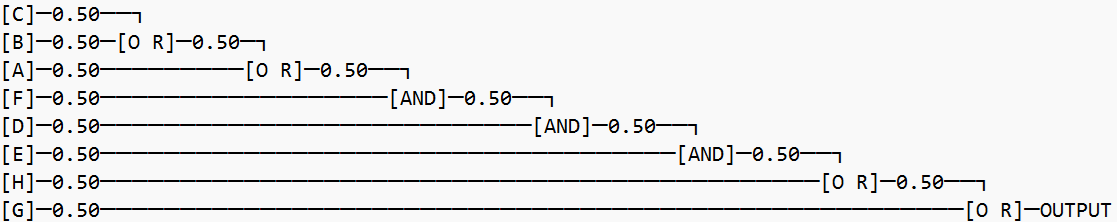}
    \caption{BACON aggregation tree for (((((((A or B) or C) and D) and E) and F) or G) or H)}
    \label{fig:bool-bacon}
\end{figure}

This expression is particularly challenging for soft logic networks due to its highly nested structure and the presence of extreme values (e.g., many variables being simultaneously 0 or 1). Such inputs create sharp decision boundaries that can be difficult for gradient-based models to approximate smoothly.
Figure ~\ref{fig:bool-bacon} presents the output of the BACON Python implementation. As shown, the program successfully recovered a fully equivalent expression, with only minor variations in the order of features.

In summary, BACON demonstrates the ability to approximate classical Boolean relationships and offers a soft, differentiable approach to modeling logic circuits. Future improvements may further enhance its performance in expressions with more complex OR-dominated structures or high-order disjunctions.

\section{BACON Iris flower classification experiment details}

The Iris flower dataset was originally introduced by Ronald Fisher in 1936 \citep{fisher1936iris} and is widely used in machine learning literature and research \citep{islr} \citep{shukla2020flower}. We used the UCI version, ID 53 \citep{uciiris} (CC-BY 4.0). Each sample in the dataset includes four numerical features: \textit{sepal length}, \textit{sepal width}, \textit{petal length}, and \textit{petal width}. These features are used to classify samples into one of three Iris species: Setosa, Versicolor, and Virginica.

We trained two BACON networks to distinguish Virginica and Setosa from the other Iris species. For Setosa, all features are reversed using \(1-x\). This is necessary in the current version because BACON logically aggregates levels of truth. The original features can be interpreted as the degree to which a feature is "large" (e.g., large petal length). However, Setosa is generally characterized by smaller feature values, so we must express statements about the degree to which a feature is small. We plan to overcome this limitation in future versions.

\begin{figure}
    \centering
    \includegraphics[width=1.0\textwidth]{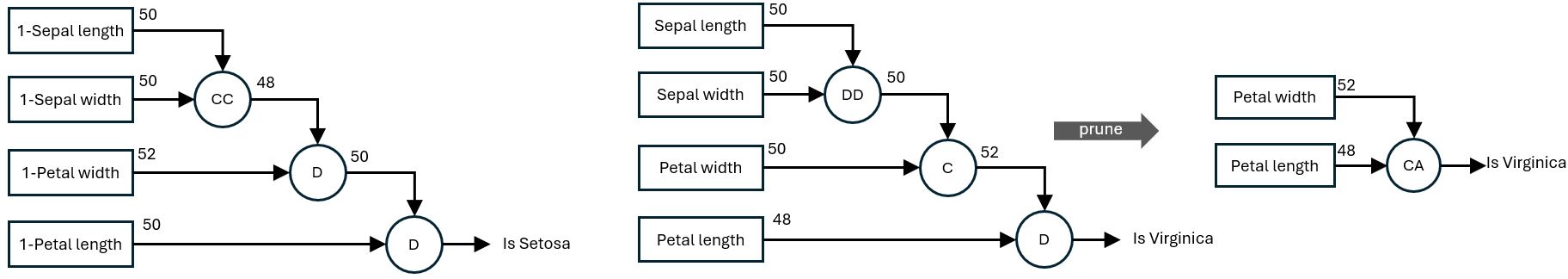}
    \caption{LSP aggregation trees for Iris classification}
    \label{fig:bacon-trees}
\end{figure}

BACON successfully generated interpretable models for Setosa and Virginica. The models produced logic-based explanations summarized in Table \ref{tab:iris}. And the corresponding LSP aggregation trees are captured by Figure \ref{fig:bacon-trees}. The absence of a Versicolor model aligns with real-world observations: Versicolor is often harder to distinguish directly, and it is more effective to eliminate Setosa and Virginica as possibilities instead.

\begin{table}
    \centering
    \caption{Iris flower classification explaination}
    \label{tab:iris}
    \begin{tabular}{llrl}
        \toprule
        \textbf{Species} & \textbf{Accuracy}  & \textbf{Explaination}\\ \midrule
        Setosa  & 96.67\%  & Sepal is small and petal is short\\ 
        Versicolor  & NA & (no apparent logic) \\ 
        Virgincia  & 96.67\% & Petal is long and wide \\
        \bottomrule
    \end{tabular}
\end{table}

In summary, BACON demonstrates strong explainability in classifying Iris flower species by constructing interpretable logic trees. It excels at clearly separating Setosa and Virginica from the rest, while its inability to generate a coherent classifier for Versicolor naturally reflects the real-world ambiguity of this class.

\section{BACON house purchasing decision experiments}

For house purchasing decision experiment, we use the US Real Estate Dataset from Kaggle \citep{ahmed_shahriar_sakib_2024}. The dataset containers 2.2 million samples with five numeric fields: \textit{price}, \textit{bed}, \textit{bath}, \textit{acre\_lot} and \textit{house\_size}. To test BACON's ability to extract decision logic our of noisy data, we created several purchase-no-purchase targets driven by logics that is hidden to BACON: 
\begin{itemize}
    \item Case A: Buy a house if it has more than 4 bedrooms, 3 baths AND is larger than 3000 square feet.
    \item Case B: Buy a house if it has more than 4 bedroom OR 3 baths OR a lot larger than 0.5 acre.
\end{itemize}

\begin{figure}
    \centering
    \includegraphics[width=.9\textwidth]{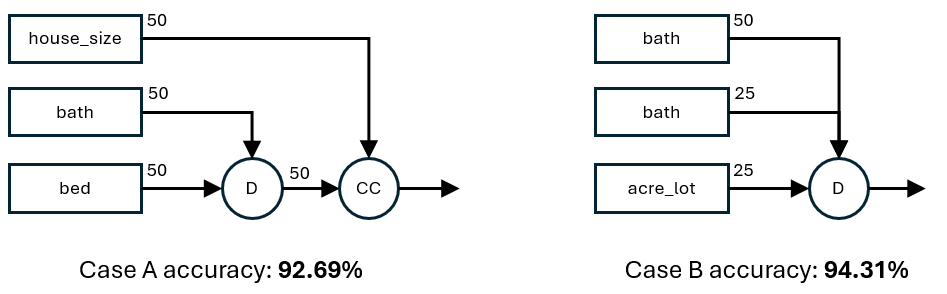}
    \caption{House purchasing decisions}
    \label{fig:houses}
\end{figure}

During training, all other numeric fields—including the \textit{zip\_code} field—serve as noise to evaluate BACON’s robustness. In all cases, BACON successfully identified the key decision-making factors and their interrelationships, while filtering out irrelevant features. Figure \ref{fig:houses} illustrates LSP aggregation trees trained with BACON for the three scenarios.

With simple boundary checks, we were able to recover approximated threshold values from the original logic, as shown in Table \ref{tab:house-prices}

\begin{table}[H]
    \centering
    \caption{House purchasing scenarios}
    \label{tab:house-prices}
    \resizebox{\textwidth}{!}{%
    \begin{tabular}{l r cc cc cc cc cc}
        \toprule
        \textbf{Scenario} & \textbf{Accuracy} & \multicolumn{2}{c}{\textbf{Bed}} & \multicolumn{2}{c}{\textbf{Bath}} & \multicolumn{2}{c}{\textbf{Lot}} & \multicolumn{2}{c}{\textbf{Size}} & \multicolumn{2}{c}{\textbf{Price}}\\
        \cmidrule(lr){3-12}
                         &                   & \textbf{act.} & \textbf{est.} & \textbf{act.} & \textbf{est.} & \textbf{act.} & \textbf{est.} & \textbf{act.} & \textbf{est.} & \textbf{act.} & \textbf{est.}\\
        \midrule
        Case A & 92.69\%  & 4.0 & 4.0 & 3.0 & 3.5 & -  & - & 3000.0 & 3454.50 & - & - \\
        Case B & 94.31\%  & 4.0  & 4.0 & 3.0 & 3.0 & 0.50 & 0.59  & - & - & - & - \\
        \bottomrule
    \end{tabular}
    }
\end{table}

\section{Andness-directed GCD with two variables}
\label{appendix:d}
The following table is trasripted from \citep{lsp2025} as a summary of GCD aggregators, which are the foundation of LSP logic modeling language:

\begin{table}[H]
\centering
\caption{Andness-directed GCD with two variables}
\resizebox{\textwidth}{!}{%
\begin{tabular}{lllllll}
\toprule
\multicolumn{5}{l}{}                                                                                    & Andess                               & Verbalization \\
\hline
\multirow{23}{*}{GCD} & \multirow{11}{*}{Conjunctive} & \multirow{8}{*}{Hard conjunction} & CC  & Drastic conjunction     & \(2\)                             & \multirow{9}{*}{"Must have all"} \\
                      &                               &                                   & HHC & High hyper-conjunction  & \([5/4,2]\)   \\
                      &                               &                                   & CP  & Product t-norm          & \(5/4\)                            \\
                      &                               &                                   & LHC & Low hyper-conjunction   & \([1,5/4]\)    \\
                      &                               &                                   & C   & Pure conjunction        & \(1\)                              \\
                      &                               &                                   & HC+ & High hard conjunction   & \(13/14\)                          \\
                      &                               &                                   & HC  & Medium hard conjunction & \(12/14\)                          \\
                      &                               &                                   & HC- & Low hard conjunction    & \(11/14\)                          \\
                      \cline{3-7}
                      &                               & \multirow{3}{*}{Soft conjunction} & SC+ & High soft conjunction   & \(10/14\)                         & \multirow{3}{*}{"Nice to have most"} \\
                      &                               &                                   & SC  & Medium soft conjunction &  \(9/14\)                           \\
                      &                               &                                   & SC- & Low soft conjunction    & \(8/14\)                           \\
                      \cline{2-7}
                      & Neutral                       & Arithmetic mean                   & A   & Logic neutrality        & \(7/14\)                          & "Nice to have" \\
                      \cline{2-7}
                      & \multirow{11}{*}{Disjunctive} & \multirow{3}{*}{Soft disjunction} & SD- & Low soft disjunction    & \(6/14\)                          & \multirow{3}{*}{"Nice to have some"} \\
                      &                               &                                   & SD  & Medium soft disjunction & \(5/14\)                           \\
                      &                               &                                   & SD+ & High soft disjunction   & \(4/14\)                           \\
                      \cline{3-7}
                      &                               & \multirow{8}{*}{Hard disjunction} & HD- & Low hard disjunction    & \(3/14\)                          & \multirow{8}{*}{"Enough to have any"} \\
                      &                               &                                   & HD  & Medium hard disjunction & \(2/14\)                           \\
                      &                               &                                   & HD+ & High hard disjunction   & \(1/14\)                           \\
                      &                               &                                   & D   & Pure disjunction        & \(0\)                              \\
                      &                               &                                   & LHD & Low hyper-disjunction   & \([-1/4,0]\)  \\
                      &                               &                                   & DP  & Product t-conorm        & \(-1/4\)                           \\
                      &                               &                                   & HHD & High hyper-disjunction  & \([-1,-1/4]\)  \\
                      &                               &                                   & DD  & Drastic disjunction     & \(-1\)  \\
                      \bottomrule
\end{tabular}
}
\end{table}

\section{BACON training and hyperparameter}
\label{appendix:e}

BACON training can be conducted on a common laptop with 16GB or memory and Python 3, plus required libraries like NumPy, PyTorch and scikit-learn. Table \ref{tab:hyperparameters} summarizes the available hyper-parametes to fine-tune BACON behaviors.

\begin{table}
    \centering
    \caption{BACON hyper-parameters}
    \label{tab:hyperparameters}
    \begin{tabularx}{\textwidth}{l X}
        \toprule
        \textbf{Parameter} & \textbf{Function} \\
        \midrule
        \textit{acceptance\_threshold} & The minimum acceptable accuracy. BACON will keep making more attempts if it has found a solution but not meeting the required performance.\\
        \textit{attempts} & Sometimes BACON searches for the best permutation may fail, as it tries only a small portion of all permutations. Although the default value is 100, in most cases the training can finish under 5 attempts. If a network fails to converge after 5 attempts, you may want to adjust other hyper-parameters and try again. \\
        \textit{freeze\_loss\_threshold} & Control the threshold to freeze the permutation layer into a one-hot assignment matrix. Default value is 0.01. Note if you use \textit{loss\_amplifier} to amplify losses, this threshold need to be multiplied by the amplifier as well.\\
        \textit{is\_frozen} & Skip feature permutation and directly use features in the given order. This allows an human expert to assert desired feature ranking. \\
        \textit{lock\_loss\_tolerance} & The maximum allowed loss degradation when the Gumbel-Sinkhorn matrix is replaced by a one-hot matrix. BACON abandons the freeze attempt if loss degradation is too high.\\
        \textit{loss\_amplifier} & Because all inputs are normalized to \([0,1]\), using an amplifier in complex cases can help to avoid vanishing gradients. Default value is 1.0 (no amplification). \\
        \textit{max\_epochs} & The maximum training epochs within an attempt.\\
        \textit{save\_model} & Whether to auto-save a model if it satisfies the \textit{acceptance\_threshold}.\\
        \textit{save\_path} & File name for the saved model. Default is "./assembler.pth".\\
        \textit{tree\_layout} & Layout of aggregation tree. Default is "left" for a left-associative binary tree. Although there's a "balanced" option available, we haven't tested this option.\\
        \textit{weight\_penalty\_strength} & How much penalty is applied to unbalanced weights. For feature importance discovery, a strong penalty (i.e. 1e-2) should be applied to encourage balanced weights. For better accuracy, this penalty can be relaxed (i.e. 1e-4) to allow more flexible, but sometimes less easily interpretable models. \\
        \bottomrule
    \end{tabularx}
\end{table}

\section{Breast cancer diagnosis experiment details}
\label{appendix:f}
\subsection{Feature normalization}
As stated in the paper, the BACON model does not require feature selection prior to training. Since BACON operates on graded logic, all input features must be normalized to the \([0,1]\) range, where each value represents a degree of truth for a corresponding claim about the feature. For example, a normalized radius value expresses the degree to which the statement “the radius is large” holds true—1 indicating the largest observed radius and 0 the smallest. In the breast cancer dataset, all numeric features are interpreted in this manner. 

Outliers can compress the range of the remaining data, making training more difficult. Therefore, they must be handled during normalization. In our experiment, we used a custom sigmoid-based normalizer that pushes extreme values toward 0 or 1 while stretching the middle range. This allows the model to better capture subtle differences among typical samples. We also tested a standard Min-Max scaler and found it resulted in slightly worse performance. While the sigmoid normalizer improves training, it's less clear whether the resulting "degree of truth" aligns with human interpretation. We argue that this interpretation should depend on the context and plan on supporting more flexible, non-uniform scaling methods in future work.

\subsection{Pruning the binary tree}

Table \ref{tab:full-tree} summarizes the original BACON network (left-associative binary tree) before pruning. This network is trained with relaxed weight constraints, prioritizing for accuracy instead of explainability, while the network presented in the paper body prioritizes explainability with more strict weight regulations. The BACON framework includes a \textit{prune\_features} method that removes a specified number of leaf nodes. As shown in the last column, most features can be pruned without affecting model accuracy, suggesting they are not essential to the decision process. In the paper text, we chose the top 5 features that the most impact on accuracy when pruned. For example, although \textit{concave\_point3} ranks high in the importance list, it is excluded because pruning it does not affect model accuracy.

\begin{table}
    \centering
    \caption{BACON breast cancer diagnosis tree before pruning}
    \label{tab:full-tree}
    \resizebox{\textwidth}{!}{%
    \begin{tabular}{rllrrrr}
        \toprule
        \textbf{Layer} & \textbf{Left Feature}  &\textbf{Right Feature}  &\textbf{w (left)} & \textbf{a (bias)} & \textbf{1-w(right)} & \textbf{Accuracy}\\
        \midrule
        1 & fractal\_dimension3 &  smoothness2 & 0.6225 & 1.7061 & 0.3775  & 98.42\%\\
        2 & Node1 & texture2 & 0.6225 & 1.3767 & 0.3775 & 98.42\%\\
        3 & Node2 & perimeter1 & 0.6225 & -0.1532 & 0.3775 & 98.42\%\\
        4 & Node3 & compactness1 & 0.6225 & 1.5872 & 0.3775 & 98.42\% \\
        5 & Node4 & smoothness1 & 0.6225 & 0.4842 & 0.3775 & 98.42\%\\
        6 & Node5 & symmetry1 & 0.6225 & 1.8122 & 0.3775 & 98.42\%\\
        7 & Node6 & fractal\_dimension2 & 0.6225 & 1.5481 & 0.3775 & 98.42\%\\
        8 & Node7 & texture1 & 0.6225 & 1.5347 & 0.3775 & 98.42\%\\
        9 & Node8 & compactness3 & 0.6225 & 1.6550 & 0.3775 & 98.42\%\\
        10 & Node9 & fractal\_dimension1 & 0.6225 & 1.5337 & 0.3775 & 98.42\% \\
        11 & Node10 & concavity2 & 0.6225 & -0.7101 & 0.3775 & 98.42\%\\
        12 & Node11 & concave\_points2 & 0.6225 & -0.6050 & 0.3775 & 98.42\%\\
        13 & Node12 & compactness2 & 0.6225 & -0.8014 & 0.3775 & 98.42\%\\
        14 & Node13 & symmetry2 & 0.6225 & 1.9969 & 0.3775 & 98.42\%\\
        15 & Node14 & concavity1 & 0.6225 & 1.9263 & 0.3775 & 98.42\%\\
        16 & Node15 & \textbf{area3} & 0.6226 & -0.7999 & 0.3774 & 98.42\%\\
        17 & Node16 & smoothness3 & 0.7365 & 1.6205 & 0.2635 & 91.04\%\\
        18 & Node17 & \textbf{concave\_points1} & 0.7979 & -0.6530 & 0.2021 & 91.04\%\\
        19 & Node18 & concave\_points3 & 0.7909 & 1.8086 & 0.2091 & 87.70\%\\
        20 & Node19 & \textbf{area2} & 0.7819 & -0.9105 & 0.2181 & 87.70\%\\
        21 & Node20 & radius2 & 0.6848 & 0.2440 & 0.3152 & 85.59\%\\
        22 & Node21 & texture3 & 0.8521 & 1.9154 & 0.1479 & 85.59\%\\
        23 & Node22 & \textbf{perimeter3} & 0.8782 & -0.8904 & 0.1218 & 85.59\%\\
        24 & Node23 & concavity3 & 0.9171 & 1.9449 & 0.0829 & 84.53\%\\
        25 & Node24 & \textbf{radius3} & 0.9096 & -0.9351 & 0.0904 & 84.53\%\\
        26 & Node25 & area1 & 0.8596 & 0.2491 & 0.1404 & 63.09\%\\
        27 & Node26 & symmetry3 & 0.8951 & 1.8458 & 0.1049 & 63.09\%\\
        28 & Node27 & perimeter2 & 0.9730 & -0.9525 & 0.0270 & 63.09\%\\
        29 & Node28 & radius1 & 0.9974 & 1.5575 & 0.0026 & 62.74\%\\
        \bottomrule
    \end{tabular}
    }
\end{table}

\subsection{Constructing the LSP aggregation tree}

\newcommand{\circledD}{%
  \mathbin{%
    \ooalign{%
      \hfil$\bigcirc$\hfil\cr
      \hfil\raisebox{0.05ex}{$\scriptstyle\mathrm{D}$}\hfil%
    }%
  }%
}

In this work, we chose the top five important features as highlighted in Table \ref{tab:full-tree} and manually constructed the LSP aggregation tree shown in Figure~\ref{fig:feature-aggregation} and applied certain simplifications. As a result, the presented LSP aggregation is not strictly a binary tree. Some simplification rules are straightforward. For example, \((A \circledD B) \circledD C\) can be simplified to \(A \circledD B \circledD C\), analogous to simplifying \((A \vee B) \vee C \) as \(A \vee B \vee C\) in classical Boolean logic. However, there are caveats in more complex cases. We plan to develop a utility to automatically generate and simplify LSP aggregation trees in future work.

\subsection{Multiple explanations}

\begin{figure}
    \centering
    \includegraphics[width=.9\textwidth]{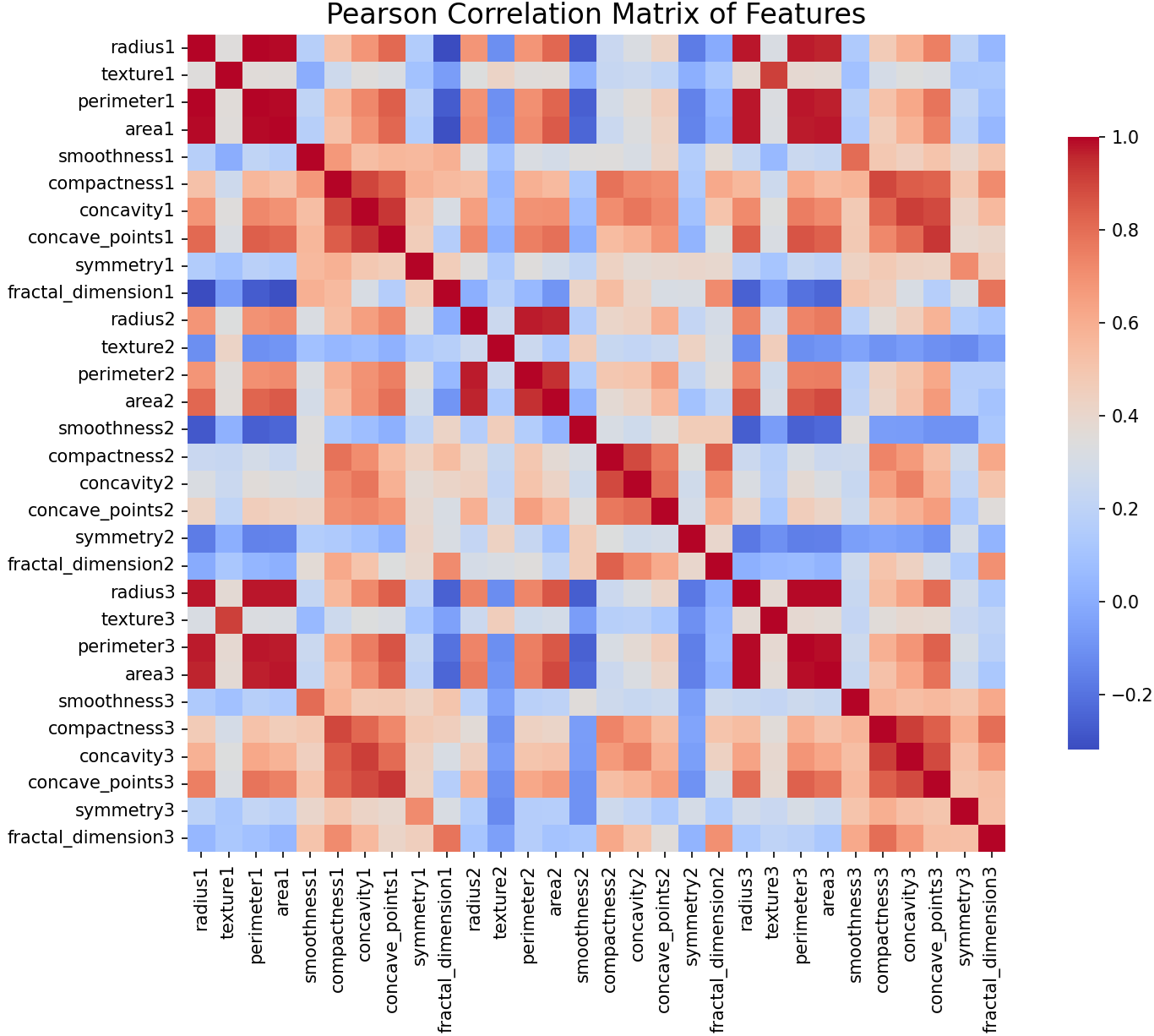}
    \caption{Breast cancer dataset feature correlation matrix}
    \label{fig:mattrix}
\end{figure}

Due to the small sample size and high correlations among features (see Figure \ref{fig:mattrix}), BACON produces multiple plausible explanations, as summarized in Table \ref{tab:multi-trees}. Nevertheless, the underlying logic remains consistent: relying on features such as large cell size and irregular shapes to detect malignant cells. We argue that these outputs are valuable to clinicians: by reviewing the generated explanations, doctors may gain insights and inspiration that support the refinement of their diagnostic practices.

In the paper, we chose the final explanation because it yielded the highest accuracy.

On the other hand, the existence of multiple plausible explanations highlights the importance of human-in-the-loop review. When an AI model is fully explainable, a human expert can examine its reasoning in detail and determine whether the explanations are relevant and trustworthy.

\begin{table}[H]
    \centering
    \caption{Multiple BACON explanations}
    \label{tab:multi-trees}
    \resizebox{\textwidth}{!}{%
    \begin{tabular}{llllllr}
        \toprule
        \textbf{Explanation} & \textbf{Top 1} & \textbf{Top 2} & \textbf{Top 3} & \textbf{Top 4} & \textbf{Top 5}  & \textbf{Accuracy}\\
        \midrule
        1 & concave\_points3 & radius2 & symmetry3  & smoothness3 & area2 & 97.36\%\\
        2 & radius2 & area2 & perimeter3 & concavity2 & compactness1 & 97.72\% \\
        3 & symmetry2 & area2 & smoothness3 & fractal\_dimension3 & concavity1 & 97.89\% \\
        4 & radius2 & perimeter3 & smoothness3 & symmetry3 & concavity1  & 97.89\% \\
        5 & area3 & smoothness3 & perimeter3 & perimeter2 & perimeter1 & 98.07\% \\
        6 & radius3 & area3 & smoothness3 & concavity1 & symmetry3  & 97.72\% \\
        7 & area3 & concave\_points3 & radius2 & texture1 & smoothness3 & 98.24\% \\
        8 & smoothness3 & concavity3 & area3 & radius1 & area2  & 98.24\% \\
        9 & radius3 & texture3 & concavity1 & concavity2 & symmetry3 & 98.42\% \\
        10 & radius3 & area3 & concavity\_points1 & area2 & perimeter3 & 98.95\% \\
        \bottomrule
    \end{tabular}
    }
\end{table}

\section{LLM generated reports}
\label{appendix:h}
The symbolic nature of LSP aggregation makes auto-generation of human-friendly reports easy. The following is an sample prompt that can be fed to a generative AI model:

\small{
\begin{verbatim}

System
=======
You are a report generator that produces a precise, human-readable, 
and insightful explanation of an LSP aggregation tree. Use your 
knowledge of LSP as well as the provided context to generate a 
clear and logical report.

Context
=======
This aggregation tree is for breast cancer diagnosis using the 
WDSD dataset.

Background
==========
LSP aggregators are defined in the following markdown table:
| GCD | Type | Subtype | Code | Name | Andness | Verbalization |
|--------|--------|--------|--------|--------|--------|--------|
...(omitted)...

Instructions
============
1. Organize the report into the following sections:
    - Overview: Summarize the overall decision logic of the tree.
    - Decision Logic Walkthrough: Explain step by step from root 
    to leaves, introducing the operator, its meaning (use the 
    verbalization from the table where applicable), and how the 
    features are combined.
    - Clinical Interpretation: Relate the aggregation logic to 
    known breast cancer diagnosis patterns (e.g., large size, 
    irregular shape), and highlight any interesting interactions 
    or compensations observed.
    
2. Use plain, human-friendly language suitable for clinicians or 
   decision-makers.

3. When describing operators, use both their code (e.g., "HHC") 
    and verbalization (e.g., "High Hyper-Conjunction — extremely 
    strict 'must have all' behavior").

4. Highlight the most influential features and explain how they 
   affect the overall decision.

5. Avoid excessive technical terms unless necessary, and prefer 
   clear analogies and explanations.

Input
======
The aggregation tree is expressed in the following JSON.
```json
{
    "operator": "HHC",
    "children": [
        {
            "feature": "concave_points3",
            "weight": 0.5,
        },
        {
            "operator": "HHD",
            "children": [
                {
                    "feature": "radius2",
                    "weight": 0.04,
                },
                {
                    "feature": "area1",
                    "weight": 0.11
                },
                {
                    "feature": "concavity1",
                    "weight": 0.34
                },
                {
                    "feature": "perimeter3",
                    "weight": 0.51
                }
            ]
        }
    ]
}
```
\end{verbatim}
}
And the following is a sample generated report:
\hrule
\textbf{{\large Overview}}

This aggregation tree provides an interpretable decision logic for breast cancer diagnosis based on the WDSD dataset. The structure of the tree reflects a highly cautious approach, ensuring that both cell irregularity and size-related factors are strictly considered. The overall decision requires strong evidence from both aspects, following an aggregation path that demands multiple conditions to be met simultaneously for a diagnosis leaning towards malignancy.

\textbf{\large Decision Logic Walkthrough}

\textbf{Top-level Aggregator: HHC — High Hyper-Conjunction (Extremely strict "must have all" behavior)}

At the root, the model uses a \textbf{High Hyper-Conjunction} (HHC), which is among the strictest conjunctive operators. This operator requires that both child nodes provide strong signals simultaneously for the decision path to indicate a malignant cell. This means that even if one path shows high risk, the decision will not be affirmative unless the other path also confirms it.

\textbf{First branch:} Feature \textit{concave\_points3} \\
This feature measures the number of concave points (sharp indentations) on the cell nucleus border. A higher value suggests increased irregularity, which is a known indicator of malignancy.

\textbf{Second branch:} 
Aggregator HHD — High Hyper-Disjunction (Extremely tolerant "enough to have any of the most extreme cases") \\
This operator is a highly permissive disjunction, meaning that the model is willing to accept any one of its child features showing extreme values as sufficient within this path.

Children of HHD:
\begin{enumerate}
    \item \textit{radius2} (Weight: 0.04) Reflects the radius of the nucleus. Lower weight suggests it is the least influential among this group.
    \item \textit{area1} (Weight: 0.11) Represents the area of the nucleus. Slightly more important but still relatively low influence.
    \item \textit{concavity1} (Weight: 0.34) Measures the degree of concave portions (dents) in the contour. Significant contributor in this group, highlighting that the degree of irregularity is critical.
    \item \textit{perimeter3} (Weight: 0.51) Captures the length of the nucleus border. Most influential feature in this group, suggesting that larger or irregular borders are strong risk indicators.
\end{enumerate}

\textbf{\large Clinical Interpretation} 

The aggregation tree reflects a clinically cautious and conservative diagnostic logic:

\textbf{The model requires both high concave points (\textit{concave\_points3}) and strong evidence from the size-related group (HHD) to consider the cell malignant.}

\textit{concave\_points3} plays a decisive role, acting as a gatekeeper that must be high regardless of what happens in the size group. The HHD branch, although permissive in combining its inputs, is treated strictly at the root level due to the HHC operator. This means that the model tolerates multiple possible manifestations of size-related abnormalities but does not allow them to override the need for high irregularity. \textit{perimeter3} and \textit{concavity1} are the standout contributors within the HHD group, confirming that larger perimeter and irregular borders are strong secondary indicators.

\textbf{\large Key clinical takeaways:}

Irregular cell shape (concave points and concavity) is mandatory for malignancy prediction.

Size indicators (perimeter, area, radius) support the diagnosis but cannot compensate for a lack of irregularity.

This aligns well with known pathology principles where malignant cells often display both abnormal size and irregular edges, but shape irregularity remains a stronger, more decisive feature.

\section{Comparison with Decision Trees}

\begin{figure}[ht]
  \centering
  \includegraphics[width=\textwidth]{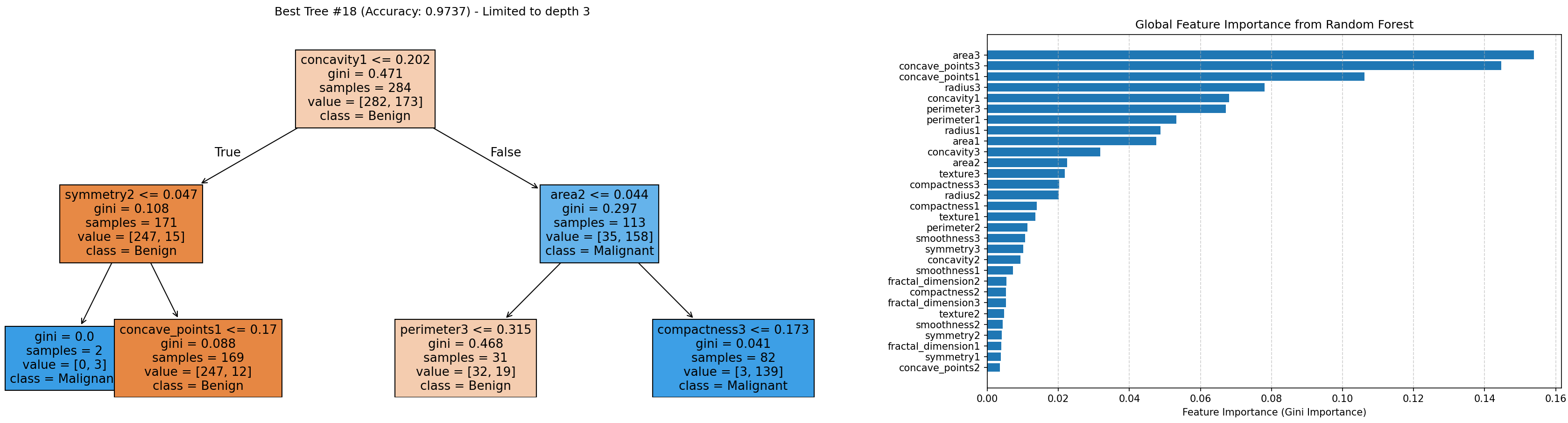}
  \caption{Decision Tree for breast cancer diagnosis}
  \label{fig:decision tree}
\end{figure}

Figure \ref{fig:decision tree} illustrates the best performing Decision Tree after training a Random Forest for breast cancer diagnosis. While the extracted decision tree achieves reasonable accuracy and captures some logical patterns such as emphasizing large area (\textit{area3}) and irregular shapes (\textit{concave\_points3}, \textit{concavity1}). Its decision logic is not aligned with common human reasoning practices. First, its decision logic is based on hard splits at specific thresholds (e.g., \textit{area2} <= 0.044), which human often operates on graded satisfaction rather than hard thresholds. Second, usage of very small thresholds can be unintuitive to humans without proper context. It is important to clarify that the concern is not the use of normalized values themselves, but rather the way these values are treated in the decision process. BACON interprets normalized values as degrees of truth and allows for graded, continuous reasoning, whereas decision trees apply rigid binary thresholds, resulting in abrupt decision boundaries. Third, the presence of features like \textit{compactness3}, which are shown to have low global importance, highlights the risk of over-interpreting local splits without considering the ensemble's overall feature prioritization.

\end{document}